\documentclass[runningheads]{llncs}
\usepackage[T1]{fontenc}
\usepackage{graphicx,verbatim}
\usepackage{xurl}
\usepackage{pifont}
\usepackage{hyperref}
\usepackage{amsmath}
\usepackage{booktabs}
\usepackage{bbding}
\usepackage{tabularx}
\usepackage{amsfonts}
\usepackage{makecell}

\usepackage{marvosym}

\begin{document}
\title{Vessel2Brain: Edge-Aware Multi-Scale Flow Matching for 3D Brain Image Generation from Vessel Masks}
\titlerunning{Vessel2Brain}
%



\author{Zhi Chen\inst{1}
\and Tianyang Zhang\inst{2}
\and Theodoros N. Arvanitis\inst{1}
\and Le Zhang\inst{1}\textsuperscript{(\Letter)}
}
\authorrunning{J. Liu et al.}
\institute{School of Engineering, College of Engineering and Physical Sciences,\\ University of Birmingham, Birmingham, UK \\
\and Institute of Biomedical Engineering, University of Oxford, Oxford, UK \\
     \email{axd1038@student.bham.ac.uk; l.zhang.16@bham.ac.uk}
}
  
\maketitle              
%

\begin{abstract}

Generating realistic 3D brain MRI volumes from sparse anatomical priors is a highly ill-posed yet clinically valuable problem, with important applications in data augmentation, simulation, and downstream neuroimaging analysis. 
In this work, we investigate the challenging task of synthesising full 3D brain MRI volumes conditioned solely on vascular label maps. 
While vascular structures provide strong anatomical and topological cues, they occupy only a small fraction of the brain volume and offer no direct information about surrounding tissue appearance.
To address this challenge, we propose Vessel2Brain, an edge-aware multi-scale conditional flow matching framework that explicitly incorporates vascular structure information into the generative process. 
Our approach formulates vessel-guided brain image synthesis as a conditional continuous flow, enabling stable training and flexible generation while preserving structural correspondence. 
A multi-scale flow matching strategy is introduced to jointly model global anatomical coherence and fine-grained local details, with auxiliary velocity predictions providing additional supervision across spatial resolutions. 
Furthermore, we design an edge-aware objective that emphasises vascular boundaries during training, encouraging anatomically faithful tissue synthesis around vessel structures.
We evaluate Vessel2Brain on two public 3D brain MRI datasets with paired vascular annotations. 
Extensive quantitative and qualitative experiments demonstrate that our method consistently outperforms state-of-the-art generative models in terms of reconstruction accuracy, structural consistency, and distributional realism. 
\textit{Our codes are available after review}.

\end{abstract}

\section{Introduction}

Three-dimensional brain MRI plays a critical role in neurological disease diagnosis, treatment planning, and population-level brain analysis.
However, acquiring well-annotated 3D brain MRI data remains expensive and time-consuming, often constrained by scanner availability, acquisition protocols, and privacy regulations. 
In contrast, vascular structures can be more readily obtained through dedicated angiographic imaging or robust automatic segmentation pipelines, making vascular label maps an attractive source of structural prior information.

Conditional medical image generation aims to model relationships between medical images under structural or modality constraints and has been widely studied. 
Early approaches were predominantly based on generative adversarial networks (GANs), which produced visually realistic results but often suffered from training instability and limited anatomical 
fidelity~\cite{zhang2019missing,zhang2019unsupervised,mahapatra2019image}.
Subsequent variational autoencoder (VAE)-based models improved robustness under one-to-many mappings, though frequently at the cost of over-smoothing and reduced structural detail~\cite{varma2025medvaeefficientautomatedinterpretation}. 
More recently, diffusion-based models have emerged as a powerful paradigm for conditional medical image generation by progressively denoising samples from simple distributions~\cite{wolleb2022diffusion}.

Below, we highlight the following two key challenges that need to be addressed for generating full 3D brain MRI volumes conditioned solely on vascular label maps:
\textit{(a) How to generate anatomically plausible 3D brain appearance under extremely sparse conditioning?}
Vascular masks provide strong topological cues but cover only a small portion of voxels and contain no direct information about surrounding tissue intensities.
However, most conditional generative methods are developed for dense conditions~\cite{Dorjsembe_2024} and may fail to robustly infer missing context when the conditioning signal is spatially limited.
\textit{(b) How to learn a reliable conditional generative model with scarce paired vessel--MRI training data?}
Paired vascular annotations and 3D brain MRI volumes are often limited due to acquisition costs, protocol differences, and annotation difficulty, which restricts supervised learning and reduces coverage of anatomical and domain variability.
In such low-data regimes, high-capacity generative models can easily overfit to the training distribution, leading to poor generalization and unstable conditional synthesis across unseen subjects or scanners.
Moreover, the sparsity of the conditioning signal further amplifies this issue.
Consequently, effective training strategies that improve robustness and data efficiency are crucial to learn transferable vessel-to-brain generation.

To address the aforementioned challenges, we propose Vessel2Brain, a conditional flow matching framework specifically designed for vessel-guided 3D brain image generation. 
By modeling the generative process as a continuous conditional flow, our approach enables stable training and flexible sampling while explicitly incorporating vascular structural priors. 
In addition, we introduce multi-scale supervision and structure-aware constraints to jointly capture global anatomical coherence and local vascular alignment.
\textbf{Our main contributions} are summarized as follows:
\textbf{(1)} We are the first to study the task of generating full 3D brain MRI volumes conditioned solely on sparse vascular label maps, establishing a challenging vessel-to-brain synthesis setting under extremely limited structural priors.
\textbf{(2)} We propose Vessel2Brain, a novel conditional flow matching framework for high-fidelity 3D brain MRI generation from vascular masks, explicitly addressing the challenges of high-dimensional generation with structure-sparse conditioning.
\textbf{(3)} Extensive experiments on two public datasets demonstrate that Vessel2Brain achieves state-of-the-art performance against representative baselines, consistently improving reconstruction accuracy, structural consistency, and distributional realism under sparse vessel conditioning.

\begin{figure}[t]
    \centering
    \includegraphics[width=\textwidth]{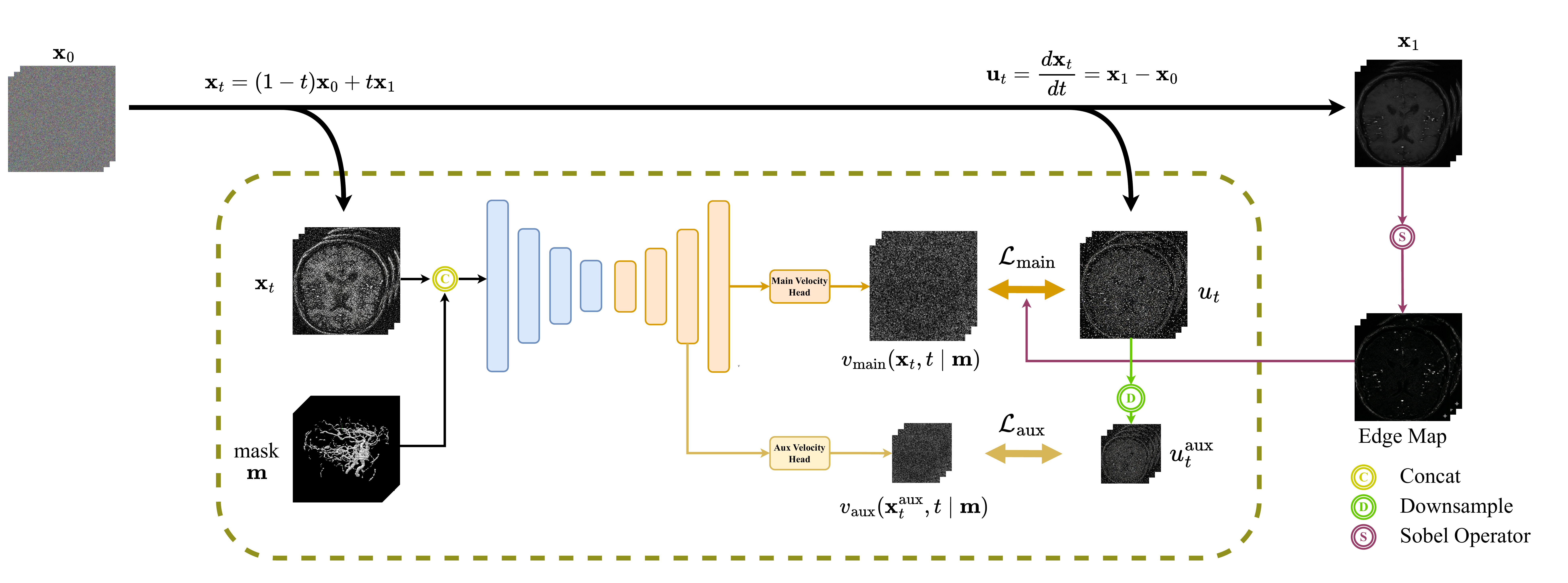}
    \caption{Overview of the proposed Vessel2Brain framework. 
    Given a 3D vascular label map and a time step $t$, a UNet-style FlowNet conditioned on timestep and label embeddings learns a velocity field for conditional flow matching.
    Velocity fields are predicted at multiple spatial resolutions through a main velocity head and an auxiliary velocity head. 
    Edge-aware weighting derived from Sobel operators on the vascular volume emphasises vessel boundaries during training. 
    The predicted velocity fields are integrated to generate the final 3D brain MRI volume, while auxiliary outputs provide additional multi-scale supervision.}
    \label{framework}
\end{figure}

\section{Methodology}
\subsection{Problem Formulation}
Let $\mathbf{m} \in \{0,1\}^{H \times W \times D}$ denote a 3D vascular label map and $\mathbf{x}\in\mathbb{R}^{H \times W \times D}$ the corresponding brain MRI volume. 
Our goal is to learn a conditional generative model that approximates $p(\mathbf{x}\mid\mathbf{m})$ and synthesises anatomically plausible brain MRIs consistent with the vessel structure.

We adopt conditional flow matching and model generation as a continuous-time transport from a simple base distribution to the target conditional distribution~\cite{lipman2023flowmatchinggenerativemodeling}.
Given a target sample $\mathbf{x}_1=\mathbf{x}\sim p(\mathbf{x}\mid\mathbf{m})$ and a noise sample $\mathbf{x}_0\sim\mathcal{N}(\mathbf{0},\mathbf{I})$, we define a linear interpolation path following the rectified flow formulation~\cite{liu2022flowstraightfastlearning}:
\begin{equation}
\mathbf{x}_t = (1-t)\mathbf{x}_0 + t\mathbf{x}_1,\quad t\sim\mathcal{U}(0,1),
\end{equation}
Notably, the velocity induced by the linear path is time-invariant. Under this linear interpolation, it can be written as:
\begin{equation}
\mathbf{u}_t = \frac{d\mathbf{x}_t}{dt} = \mathbf{x}_1-\mathbf{x}_0.
\end{equation}
We train a time-dependent conditional velocity field $v_\theta(\mathbf{x}_t,t\mid\mathbf{m})$ to match $\mathbf{u}_t$ at random timesteps $t$.
At inference, integrating $v_\theta$ from $t=0$ to $t=1$ transforms Gaussian noise into a synthetic brain MRI conditioned on vascular map $\mathbf{m}$. Compared with diffusion-based formulations that rely on iterative noise injection and denoising, conditional flow matching directly learns a deterministic velocity field along a predefined interpolation path, which improves training stability and reduces sampling complexity under sparse structural conditioning.

\subsection{UNet-style conditional FlowNet}

The overall architecture of Vessel2Brain is illustrated in Fig.~\ref{framework}.
Our framework adopts a UNet-style FlowNet to parameterize the conditional velocity field
$v_\theta(\mathbf{x}_t,t\mid\mathbf{m})$ for vessel-guided 3D brain MRI generation.

\textbf{Architecture Overview.}
We employ a UNet-style~\cite{ronneberger2015unetconvolutionalnetworksbiomedical} encoder--decoder architecture to model conditional flow matching.
The vascular label map $\mathbf{m}$ is incorporated as an explicit conditioning signal by
concatenating it with the intermediate flow state, ensuring that vascular structural
information is available throughout the network.
In addition, a continuous timestep embedding is computed according to the flow matching
formulation and injected into all residual blocks via Feature-wise Linear Modulation~\cite{perez2017filmvisualreasoninggeneral} (FiLM).
Specifically, the timestep embedding is transformed into a learned bias term $\beta(t)$,
which modulates intermediate features as $\mathbf{h}'=\mathbf{h}+\beta(t)$.
This design enables the network to adapt its predictions across different stages of the
flow trajectory while remaining explicitly conditioned on vascular structure.
The backbone consists of four resolution levels with channel widths
$\{32,64,128,256\}$.
Each encoder stage is composed of time-conditioned residual blocks and attention layers
followed by spatial downsampling.
The decoder mirrors this structure with symmetric upsampling operations and skip
connections to preserve fine-grained spatial details.
A middle block with residual and attention layers further captures global anatomical
context.
Together, this architecture allows the model to jointly leverage sparse vascular priors,
temporal flow information, and multi-scale spatial representations for anatomically
consistent 3D brain MRI synthesis.

\textbf{Auxiliary Velocity Prediction.}
To improve training stability and capture global anatomical structure in early stages,
the FlowNet introduces an auxiliary velocity prediction at a coarser spatial resolution.
Specifically, besides the main full-resolution output
$v_{\text{main}}(\mathbf{x}_t,t\mid\mathbf{m})$, we form a downsampled flow state:
\begin{equation}
\mathbf{x}^{\text{aux}}_t=\mathcal{D}(\mathbf{x}_t),
\end{equation}
where $\mathcal{D}(\cdot)$ denotes a spatial downsampling operator, and attach an
auxiliary velocity head to an intermediate decoder feature map to predict
$v_{\text{aux}}(\mathbf{x}^{\text{aux}}_t,t\mid\mathbf{m})$.
Both velocity heads consist of Group Normalisation, SiLU activation, and 3D convolution
layers.
The auxiliary prediction provides complementary coarse-scale supervision, encouraging
the model to first capture long-range anatomical context, while the main branch refines
high-frequency tissue details at full resolution.

The auxiliary head is trained using a standard L1 flow matching objective:
\begin{equation}
\mathcal{L}_{\text{aux}}
=
\mathbb{E}\!\left[
\big\|
v_{\text{aux}}(\mathbf{x}_t^{\text{aux}}, t \mid \mathbf{m})
-
u_t^{\text{aux}}
\big\|_1
\right],
\end{equation}
where $u_t^{\text{aux}}$ is obtained by downsampling the target velocity
$u_t=\mathbf{x}_1-\mathbf{x}_0$ to the auxiliary resolution.
This auxiliary supervision stabilises optimisation and promotes consistent global
structure learning across scales.

\textbf{Edge-Weighted Flow Matching Loss.}
Following the formulation of conditional flow matching, the network is trained to match the target velocity
$u_t=\mathbf{x}_1-\mathbf{x}_0$ along the interpolation path
$\mathbf{x}_t=(1-t)\mathbf{x}_0+t\mathbf{x}_1$, where
$\mathbf{x}_0\sim\mathcal{N}(0,I)$ and $t\sim\mathcal{U}(0,1)$.
To emphasise anatomical boundaries and fine vascular-aligned structures, we compute a
3D edge magnitude map from the target volume $\mathbf{x}_1$ using Sobel operators~\cite{sobel19683x3} and
normalise it by its spatial mean to obtain an edge weighting map $w(\mathbf{x}_1)$.
The main velocity head is supervised using an edge-aware L1 flow matching loss:
\begin{equation}
\mathcal{L}_{\text{main}}
=\mathbb{E}_{t,\mathbf{x}_0,\mathbf{x}_1}\!\left[
\Big((1-\lambda_{\text{edge}})+\lambda_{\text{edge}}\,w(\mathbf{x}_1)\Big)\,
\big\|v_{\text{main}}(\mathbf{x}_t,t\mid\mathbf{m})-u_t\big\|_1
\right].
\end{equation}

\textbf{Overall Training Objective.}
The final training objective combines the edge-weighted main flow matching loss and the auxiliary coarse-scale loss:
\begin{equation}
\mathcal{L}
=
\lambda_{\text{main}}\,\mathcal{L}_{\text{main}}
+
\lambda_{\text{aux}}\,\mathcal{L}_{\text{aux}},
\end{equation}
where $\lambda_{\text{main}}$ and $\lambda_{\text{aux}}$ control the relative contribution of each term.
In practice, $\mathcal{L}_{\text{main}}$ serves as the primary supervision at full resolution and encourages high-fidelity tissue appearance while sharpening vessel-aligned boundaries through edge weighting.
Meanwhile, $\mathcal{L}_{\text{aux}}$ provides complementary multi-scale guidance by constraining the transport velocity at a coarser resolution.
We set $\lambda_{\text{main}}$ as the dominant weight and use a smaller $\lambda_{\text{aux}}$ to regularise training without over-constraining fine details.

\section{Experiments}


\textbf{Datasets and Augmentation.}
Experiments were conducted on two publicly available datasets, \emph{CereVessMRA}~\cite{guo2024self} and \emph{TopBrain Challenge dataset}~\cite{topcowchallenge}, both of which provide paired 3D brain MRI volumes and corresponding vascular annotations.
All MRI volumes and vascular masks were resampled to a unified spatial resolution of $128 \times 128 \times 128$ using SimpleITK~\cite{beare2018image}.
Linear interpolation was applied to MRI images, whereas nearest-neighbour interpolation was used for vascular masks to preserve discrete boundaries.
MRI intensities were normalised to the $[-1,1]$ range via Min--Max normalisation.
Vascular masks were converted into two-channel one-hot representations corresponding to background and vessel structures.
The datasets were split into training, validation, and test sets at the subject level with a ratio of 8:1:1 to prevent data leakage.
All quantitative evaluations were performed on the held-out test set.

To increase anatomical diversity and improve generalisation~\cite{cciccek20163d}, we applied a 3D data augmentation pipeline during training.
All augmentations were performed jointly on MRI volumes and vascular masks using a shared affine transformation to preserve spatial correspondence.
Every training sample was randomly augmented once by sampling a transformation from the augmentation space.
The augmentation strategy included:
\textbf{(1)} random rotations up to $15^\circ$ along all three anatomical axes,
\textbf{(2)} in-plane translations up to $1\%$ of the image dimensions,
\textbf{(3)} random horizontal and vertical flips.
All augmentations were applied on-the-fly during training and were disabled during validation and testing.

\begin{figure}[t]
    \centering
    \includegraphics[width=\textwidth,height=\textheight,keepaspectratio]{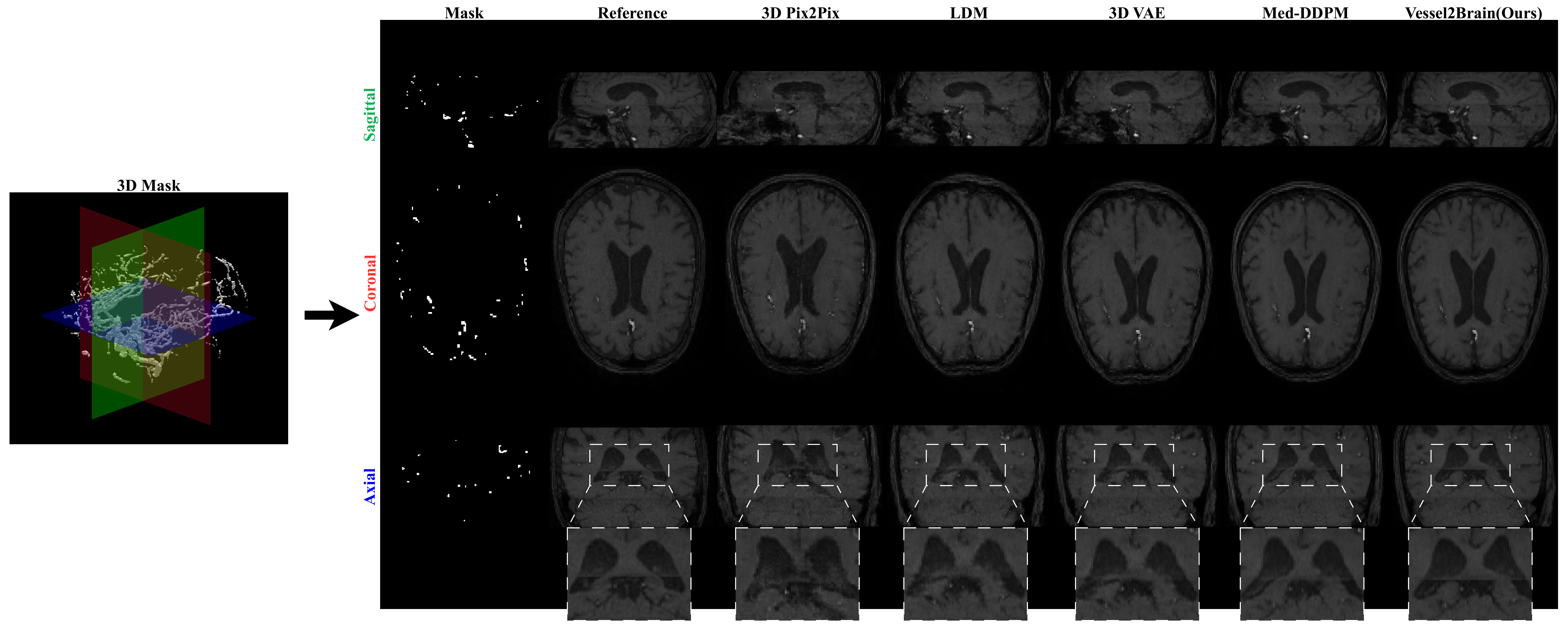}
    \caption{Qualitative comparison on the CereVessMRA dataset.
    From left to right: vascular mask, ground-truth brain MRI, 3D Pix2Pix, LDM, 3D VAE,
    Med-DDPM, and the proposed Vessel2Brain.
    Results are visualized on representative sagittal, coronal, and axial slices.
    The highlighted regions and corresponding zoomed-in views illustrate fine-grained
    anatomical details.
    Compared with baseline methods, Vessel2Brain generates more anatomically consistent
    brain structures while better preserving vascular-aligned tissue appearance under
    sparse vessel conditioning.}
    \label{CereVessMRA vis}
\end{figure}

\textbf{Implementation Details.}
All models were implemented in PyTorch Lightning and trained for 500 epochs with mixed-precision on an NVIDIA A100 GPU with 40GB of VRAM.
We used a batch size of 1 and optimized the network with AdamW using a learning rate of $3\times10^{-5}$ and a weight decay of $1\times10^{-4}$.
We set $\lambda_{\text{main}}$ to 1.0 and $\lambda_{\text{aux}}$ to 0.1 to weight the main and auxiliary losses, respectively.
To emphasize anatomical boundaries, we applied Sobel-based edge magnitude weighting to the main loss and set $\lambda_{\text{edge}}$ to 0.3.
During training, we maintained an exponential moving average of model parameters with a decay factor of 0.999.

\textbf{Evaluation Metrics and Baselines.}
We evaluate model performance using four standard image similarity metrics:
Structural Similarity Index (SSIM)~\cite{wang2004image},
Peak Signal-to-Noise Ratio (PSNR),
Mean Absolute Error (MAE), and
Normalized Mutual Information (NMI)~\cite{studholme1999overlap}.
Higher values of SSIM, PSNR, and NMI, and a lower value of MAE, indicate better image quality and closer alignment with the ground truth.
For comparison, we consider representative conditional generative baselines,
including the GAN-based 3D Pix2Pix~\cite{isola2018imagetoimagetranslationconditionaladversarial},
the VAE-based 3D Medical VAE~\cite{kapoor2023multiscalemetamorphicvae3d},
the latent diffusion model (LDM~\cite{rombach2022highresolutionimagesynthesislatent}),
and the diffusion-based Med-DDPM~\cite{Dorjsembe_2024},
covering major paradigms in medical image synthesis.


\begin{table}[t]
    \centering
    \setlength{\tabcolsep}{2.2pt}
    \renewcommand{\arraystretch}{1.3}
    \caption{Quantitative comparison on the CereVessMRA dataset.
    Results are reported as mean $\pm$ standard deviation.
    $\uparrow$/$\downarrow$ denote that higher/lower values indicate better performance.
    Vessel2Brain consistently outperforms representative GAN-, VAE-, and diffusion-based baselines across all metrics,
    highlighting its effectiveness for vessel-guided 3D brain MRI generation under sparse conditioning.}
    \begin{tabular}{l|llll}
        \hline
        \textbf{Method} & SSIM$(\%)$$\uparrow$  & PSNR$\uparrow$   & NMI$\uparrow$   & MAE$(\%)$$\downarrow$\\
        \hline
        3D Pix2Pix~\cite{isola2018imagetoimagetranslationconditionaladversarial} & $57.69(\pm8.13)$ & $23.74(\pm2.82)$ & $1.138(\pm0.017)$ & $4.57(\pm0.78)$\\
        3D Medical VAE~\cite{kapoor2023multiscalemetamorphicvae3d} & $63.17(\pm8.70)$ & $26.40(\pm2.10)$ & $1.133(\pm0.020)$ & $3.13(\pm0.89)$\\
        LDM~\cite{rombach2022highresolutionimagesynthesislatent} & $61.83(\pm9.46)$ & $26.16(\pm2.91)$ & $1.129(\pm0.018)$ & $3.23(\pm0.97)$\\
        Med-DDPM~\cite{Dorjsembe_2024} & $63.81(\pm8.57)$ & $26.79(\pm2.59)$ & $1.142    (\pm0.018)$ & $2.99(\pm1.09)$\\
        Vessel2Brain(w/o Aug.)& $63.99(\pm9.02)$ & $26.40(\pm2.72)$ & $1.139(\pm0.018)$ & $3.28(\pm0.89)$\\
        Vessel2Brain(w/o Aux.)& $64.83(\pm8.32)$ & $26.53(\pm2.31)$ & $1.138(\pm0.017)$ & $3.41(\pm0.91)$\\
        \textbf{Vessel2Brain} & $\mathbf{65.36(\pm7.36)}$ & $\mathbf{26.84(\pm2.63)}$ & $\mathbf{1.145   (\pm0.017)}$ & $\mathbf{2.97(\pm0.87)}$\\
        \hline
    \end{tabular}
    \label{CereVessMRA result}
\end{table}

\textbf{Evaluation on CereVessMRA dataset.}
Table~\ref{CereVessMRA result} reports quantitative comparisons with representative conditional generative baselines.
Overall, Vessel2Brain achieves the best performance across all metrics, attaining the highest SSIM, NMI, and PSNR, as well as the lowest MAE.
Compared with diffusion-based baselines, Vessel2Brain improves SSIM over Med-DDPM while maintaining lower MAE, indicating superior structural fidelity and reconstruction accuracy.
Notably, GAN-based 3D Pix2Pix performs substantially worse on this sparse-conditioning task, suggesting that adversarial training alone is insufficient to recover anatomically plausible brain appearance from vessel masks.
The qualitative results are shown in Fig.~\ref{CereVessMRA vis}, where the images produced by Vessel2Brain is visually more similar to the reference images than those generated by the other methods, aligning well with the quantitative results.

\begin{figure}[t]
    \centering
    \includegraphics[width=\textwidth,height=\textheight,keepaspectratio]{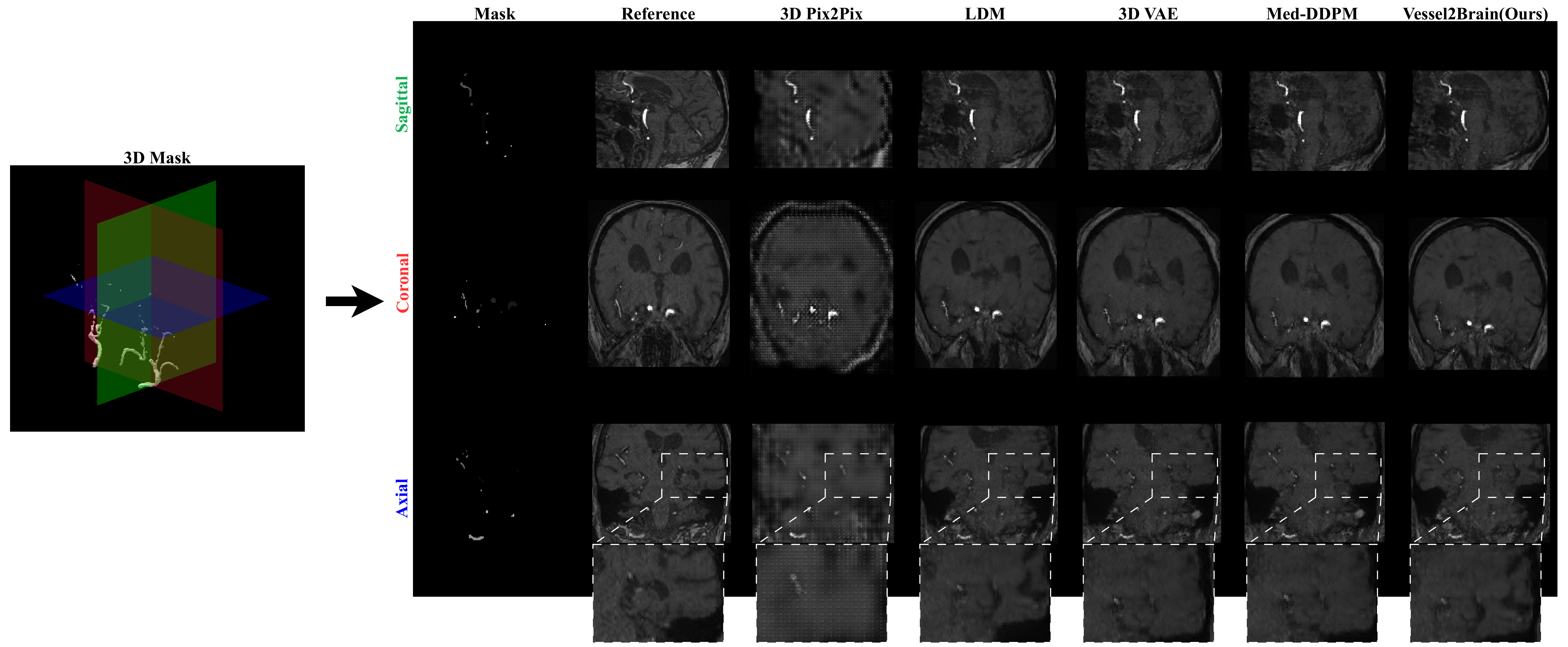}
    \caption{Qualitative comparison on the TopBrain Challenge dataset.
    The visualization setting and column order are the same as in Fig.~\ref{CereVessMRA vis}.
    Vessel2Brain produces more anatomically coherent brain structures and clearer local details than competing methods under sparse vascular conditioning.}
    \label{TopBrain vis}
\end{figure}

\begin{table}[t]
\setlength{\tabcolsep}{2.2pt}
\renewcommand{\arraystretch}{1.3}
    \centering
    \caption{Quantitative comparison on the TopBrain Challenge dataset.
    The evaluation protocol, metric definitions ($\uparrow$/$\downarrow$), and reporting format (mean $\pm$ std.) are the same as in Table~\ref{CereVessMRA result}.
    Vessel2Brain achieves the best performance across all metrics, indicating strong generalization under increased dataset heterogeneity.}
    \begin{tabular}{l|llll}
        \hline
        \textbf{Method} & SSIM$(\%)$$\uparrow$  & PSNR$\uparrow$   & NMI$\uparrow$   & MAE$(\%)$$\downarrow$\\
        \hline
        3D Pix2Pix~\cite{isola2018imagetoimagetranslationconditionaladversarial} & $17.41(\pm3.54)$ & $13.78(\pm1.19)$ & $1.052(\pm0.016)$ & $15.56(\pm3.40)$\\
        3D Medical VAE~\cite{kapoor2023multiscalemetamorphicvae3d}  & $59.40(\pm2.45)$ & $26.61(\pm0.97)$ & $1.071(\pm0.014)$ &$3.12(\pm0.30)$ \\
        LDM~\cite{rombach2022highresolutionimagesynthesislatent} & $60.97(\pm2.72)$ & $27.15(\pm0.87)$ & $1.074(\pm0.011)$ & $2.94(\pm0.33)$\\
        Med-DDPM~\cite{Dorjsembe_2024} & $59.70(\pm2.45)$ & $26.64(\pm0.69)$ & $1.072    (\pm0.008)$ & $3.11(\pm0.28)$\\
        Vessel2Brain(w/o Aug.) & $56.72(\pm2.74)$ & $25.72(\pm1.32)$ & $1.066(\pm0.015)$ & $3.07(\pm0.42)$\\
        Vessel2Brain(w/o Aux.) & $58.92(\pm2.85)$ & $26.81(\pm1.03)$ & $1.071(\pm0.017)$ & $2.97(\pm0.44)$\\
        \textbf{Vessel2Brain} & $\mathbf{60.99(\pm 2.86)}$ & $\mathbf{27.19(\pm0.93)}$ & 
        $\mathbf{1.076(\pm0.011)}$& $\mathbf{2.93(\pm0.36)}$\\
        \hline
    \end{tabular}
    \label{TopBrain result}
\end{table}

\textbf{Evaluation on TopBrain Challenge dataset.}
On the more heterogeneous TopBrain dataset (Table~\ref{TopBrain result}), Vessel2Brain again delivers the best overall performance.
It outperforms strong latent diffusion and DDPM baselines.
In particular, 3D Pix2Pix collapses on this dataset, highlighting the difficulty of learning robust vessel-to-brain synthesis under domain variability without explicit structure-preserving constraints.
The consistent gains across both datasets demonstrate that the proposed Vessel2Brain generalizes well and preserves structural correspondence even when the conditioning signal is extremely sparse.
Qualitative results are illustrated in Fig.~\ref{TopBrain vis}, the images generated by our model is visually most similar to the reference image.

\textbf{Ablation Study.}
We conduct ablation studies on both datasets to evaluate the contribution of individual components
(Tables~\ref{CereVessMRA result} and~\ref{TopBrain result}).
Removing data augmentation (w/o Aug.) leads to a clear performance drop, highlighting its importance
for robustness to dataset heterogeneity.
Similarly, excluding the auxiliary multi-scale branch (w/o Aux.) consistently degrades performance,
demonstrating the benefit of multi-scale supervision for accurate local structure modeling.
Notably, the performance degradation is reflected across multiple metrics (SSIM, PSNR, and MAE),
indicating that both global structural consistency and fine-grained appearance details are affected.
Overall, these results confirm that the full Vessel2Brain design is essential for achieving optimal
reconstruction accuracy and anatomical fidelity.

\section{Conclusion}
In this work, we introduced Vessel2Brain, an edge-aware multi-scale conditional flow matching framework for generating 3D brain MRI volumes from sparse vascular label maps. Unlike prior generative methods that rely on dense conditioning or struggle to preserve anatomical structure, Vessel2Brain explicitly models continuous velocity fields and incorporates multi-scale supervision to maintain global anatomical coherence and local structural fidelity. Experiments on two public 3D brain MRI datasets demonstrate superior reconstruction accuracy, structural consistency, and distributional realism, highlighting the potential of Vessel2Brain for data augmentation and vessel-guided brain imaging applications. In future work, we plan to extend the framework to multi-contrast or multi-domain settings, investigate stronger anatomical priors and uncertainty-aware sampling to better reflect the inherent one-to-many mapping, and explore scaling strategies for improved generalisation under limited paired training data.

    



%
%
%
\bibliographystyle{splncs04}
\bibliography{ref}
\end{document}